%% file: ActionModelVI2027.tex
\documentclass[letterpaper]{article} 
\usepackage[preprint]{aaai2027}  
\usepackage[hyphens]{url}  
\usepackage{graphicx} 
\usepackage{natbib}  
\usepackage{caption} 
\usepackage{algorithm}
\usepackage{algorithmic}

\usepackage{newfloat}
\usepackage{listings}
\DeclareCaptionStyle{ruled}{labelfont=normalfont,labelsep=colon,strut=off} 
\floatstyle{ruled}
\newfloat{listing}{tb}{lst}{}
\floatname{listing}{Listing}

\usepackage{booktabs}
\usepackage{enumerate}[inline]

\usepackage{booktabs}
\usepackage{multirow}
\usepackage{amsmath}
\usepackage{amssymb}
\usepackage{tikz}
\usetikzlibrary{positioning,arrows.meta,calc}

\input{macros}

\newcommand{\rosamei}{\textsc{rosame-i}}
\newcommand{\nesymm}{\textsc{NeSyMM}}
\newcommand{\nesyam}{\textsc{NeSyAM}}
\newcommand{\pam}{\textsc{pam}}
\newcommand{\act}[1]{\textit{#1}}
\newcommand{\role}[1]{\textsc{#1}}

\title{Neurosymbolic Action Model Learning under Partial Observability}
\author {
    Adem Kikaj\textsuperscript{\rm 1}\corresponding,
    Lennert De Smet\textsuperscript{\rm 1},
    Giuseppe Marra\textsuperscript{\rm 1},
    Luc De Raedt\textsuperscript{\rm 1}
}
\affiliations {
    \textsuperscript{\rm 1} KU Leuven, Belgium \\
    firstname.lastname@kuleuven.be
}

\begin{document}

\maketitle

\begin{abstract}
    AI planning studies how an agent can reach a goal by executing a sequence of actions.
    To plan correctly, the agent needs an action model describing when each action can be executed and how it changes the world.
    Constructing such models by hand requires domain expertise, and can be costly and error-prone.
    Action models can instead be learned from available data using existing neurosymbolic approaches, but they currently assume access to complete traces of fully observable images
    .
    These approaches fail to learn action models under partial observability where some of the images might not be present or are not fully informative of the current state of the world.
    Hence, this paper proposes \nesyam{}, a novel neurosymbolic modeling paradigm for action model learning under partial observability.
    In addition, the paper presents a unified variational framework for theoretically analysing the limitations of existing methods compared to our proposed approach.
    \nesyam{} is then tested extensively on six visual planning domains and three observation regimes to show it consistently recovers relevant parts of the true action model under partial observability.
\end{abstract}


\section{Introduction}
Given an initial state and a goal, AI planning searches for a sequence of actions that transforms the initial state into a state satisfying the goal~\cite{ghallab2004automated}.
This search relies on symbolic action models: descriptions of when actions are applicable and how they change the world, usually specified as preconditions and effects over symbolic predicates.
In \textsc{Blocksworld}, for example, picking up a block requires the block to be clear and on the table (preconditions), and causes the robot to hold it (effects).
Such models are traditionally provided by domain experts, but their manual construction is costly, error-prone, and difficult to scale.
This has motivated a long line of work on learning action models from traces~\cite{pasula2007learning,yang2007learning,zhuo2013action,aineto2019learning,lamanna2024action,aineto2024action,lamanna2025lifted}
.

Learning from symbolic traces however still requires a fully symbolic description of the traces provided by domain experts.
Consequently, more recent work considers learning from visual traces~\cite{xi2024neuro, xi2026learning} where the learner observes traces of images and executed actions together with final-state information.
Learning from such traces is a neurosymbolic problem: neural perception is required to interpret the images, while symbolic reasoning is required to recover the preconditions and effects of actions.
Unfortunately, existing neurosymbolic learning approaches make the strong assumption that visual traces are complete and fully observable.
They can not be applied to settings with missing images or when images do not contain all information of the state.
For example, if a robot's gripper is outside of its camera view, the images recorded when performing the action \act{pick-up(A)} may not show that the robot is holding block~$A$.
The images may therefore provide insufficient evidence for $holding(A)$, even though earlier observations, later observations, or the final-state description may constrain it.


In this paper, we focus on contributing to neurosymbolic action model learning under partial observability.
We start by
\begin{enumerate*}[label=\textbf{(\arabic*)}]
    \item proposing \nesyam, a novel structured approximation to the partially observable action model learning problem.
    This approximation is then
    \item theoretically compared to existing approaches by casting action model learning within a unifying variational perspective (Section~\ref{sec:vi-perspective}).
    Specifically, this perspective shows that the state of the art uses a type of mean-field approximation to the action model learning problem that relies on complete and fully observable visual traces, while our structured approximation does not.
    Finally, we
    \item compare \nesyam{} with \rosamei{} across six visual planning domains and three observation regimes. Under full observability, \nesyam{} recovers all relevant action roles, while \rosamei{} achieves higher full-candidate recovery. Under occlusion, \nesyam{} preserves relevant-role recovery more consistently, although the full-candidate comparison remains mixed.
\end{enumerate*}

\section{Preliminaries}
In all that follows, uppercase characters $X$ will denote random variables or random vectors with probability distribution $p(X)$, while lowercase characters $x$ denote instances of those variables with corresponding probabilities $p^X(x)$, or simply $p(x)$ if the distribution is clear from context.

\paragraph{Planning domains and action models.}
We consider a lifted planning domain
$
D=\langle \types, \predicates, \actions,\mathrm{AM}\rangle,
$
where \types is a set of types, \predicates is a set of predicate symbols, \actions is a set of action schemas, and $\mathrm{AM}$ is an action model.
A classical action model assigns every action schema $a(\vec{x})\in \actions$ a tuple
\[
\mathrm{AM}(a(\vec{x}))
=
\left\langle
\mathrm{Pre}(a(\vec{x})),
\mathrm{Add}(a(\vec{x})),
\mathrm{Del}(a(\vec{x}))
\right\rangle,
\]
containing its preconditions, add effects, and delete effects.

For a planning instance $I$ with a finite set of objects \objects, grounding the variables $\vec{x}$ in the action schemas $a(\vec{x})$ and predicate templates $p(\vec{x})$ with respect to \objects produces a finite set of propositions $\predicates_I$ and grounded actions $\actions_I$.
For an action schema $a(\vec{x}) \in \actions$ and a type-compatible tuple of objects $\vec{o} \in \objects$, we write $a(\vec{o})\in \actions_I$ for the corresponding grounded action.
For example, $holding$ is a predicate symbol with template $holding(x)$ that is ground into the proposition $holding(A)$, while the action schema \act{pick-up(x)} is ground into the action \act{pick-up(A)}.
A symbolic state $s \subseteq \predicates_I$ contains the grounded propositions that are true.
We write $s_p \in \{0,1\}$ for the truth value of proposition $p \in \predicates_I$, with $s_p=1$ if and only if $p\in s$.

We use mostly standard STRIPS semantics to define action applicability and state transitions \cite{fikes1971strips,ghallab2004automated}. A grounded action $a$ is applicable in state $s$ when
$
\mathrm{Pre}(a)\subseteq s.
$
Applying an applicable action produces
$
\operatorname{res}(s,a)
=
\bigl(s\setminus\mathrm{Del}(a)\bigr)
\cup\mathrm{Add}(a).
$
Propositions that are not deleted persist.
Preconditions that are not deleted remain true and are called \emph{prevail conditions} \cite{backstrom1995complexity}.
Standard STRIPS permits adding an already true proposition. 
However, our model will apply the additional constraint that an add effect can only apply to a false proposition. This modeling choice is intended to reduce reasoning shortcuts \cite{marconato2023not}.

In \textsc{Blocksworld}, the action schema \act{pick-up($x$)} illustrates these definitions. For block~$A$, the grounded action \act{pick-up(A)} requires $\mathit{clear}(A)$, $\mathit{onTable}(A)$, and $\mathit{handEmpty}$. It adds $\mathit{holding}(A)$ and deletes $\mathit{clear}(A)$, $\mathit{onTable}(A)$, and $\mathit{handEmpty}$. Therefore, considering only these relevant propositions, applying \act{pick-up(A)} transforms
$
\{
\mathit{clear}(A),
\mathit{onTable}(A),
\mathit{handEmpty}
\}
$
into
$
\{
\mathit{holding}(A)
\}.
$
In our learning setting, these precondition and effect relations are not given and must be recovered from visual traces.

\paragraph{Visual traces.}
The learner receives visual traces rather than symbolic state trajectories.
A trace has the form
$
\tau
=
(z_1,a_1,z_2,a_2,\ldots,z_{T}, a_{T}, f),
$
where $z_t$ are images, $a_t\in \actions_I$ are observed grounded actions, and $f$ is a fully observed symbolic description of the state reached after the final action.
The trace contains no image of this final state.
More formally, let $s_t\subseteq \predicates_I$ denote the symbolic state underlying image $z_t$, and let $s_{T + 1}$ denote the state reached after the final action.
Then, the sequence
$
\statevec = (s_1,\ldots,s_{T + 1})
$
forms the symbolic execution of which the first $T$ entries $\statevec_{1:T}$ are hidden.
The final symbolic state $s_{T + 1}$ is fully described by the observed description $f$.
Other observed entities are the images $\emissionvec = (z_1, \dots, z_{T})$  and grounded actions $\actionvec = (a_1, \dots, a_{T})$.
The image observations \emissionvec might in our case be partial, i.e. an image $z_t$ does not fully describe the content of its associated symbolic state $s_t$.

For example, for $t<T + 1$, a \textsc{Blocksworld} trace may contain an image $z_t$ showing block~$A$ clear and on the table, followed by \act{pick-up(A)} and a successor image $z_{t+1}$. The underlying transition changes $\mathit{holding}(A)$ from false to true and removes $\mathit{onTable}(A)$. Under partial observability, the gripper or block~$A$ may be hidden in $z_{t+1}$, so the image alone does not determine whether $\mathit{holding}(A)$ became true. Other observations, executed actions, and the final-state description may nevertheless constrain its value.

\paragraph{Probabilistic action models.}
We represent the unknown action model $\mathrm{AM}$ using a probabilistic action model (\pam{})~\cite{xi2024neuro}.
A \pam{} models the effects of actions on predicates in a lifted way by considering lifted combinations
$
c
=
\bigl(
a(\vec{x}),
p(\vec{y})
\bigr),
$
where $a(\vec{x})\in \actions$ is an action schema and $p(\vec{y})$ a type-compatible lifted predicate template for each $p \in \predicates$.
The \pam{} predicts probabilities for four possible roles
$
\role{not-involved},
\role{pre},
\role{add},
\role{del},
$
for each lifted combination $c$.
The role \role{not-involved} means that the predicate is unrelated to the action.
The role \role{pre} denotes a prevail condition, \role{add} denotes an add effect and \role{del} denotes a predicate that is both a precondition and a delete effect.
Specifically, these four mutually exclusive roles form the domain of a categorical random variable $R_p$ over the roles of predicate template $p(\vec{y})$ with conditional probability distribution $p_{\params_{\actionmodel}}(R_p \mid a)$ given an action schema $a$.
The \pam{} distribution $p_{\params_{\actionmodel}}(R_p \mid a)$ is modeled by four parameters
$
\theta_c = (p_{\role{N}}, p_{\role{P}}, p_{\role{A}}, p_{\role{D}})
$
\text{with}
$
p_{\role{N}} + p_{\role{P}} + p_{\role{A}} + p_{\role{D}} = 1,
$
for each lifted combination $c \in \mathcal{C} = \actions\times \predicates$ representing the unknown precondition and effect relations in $\mathrm{AM}(a(\vec{x}))$.
We collect the global action-model parameters as
$
\params_{\mathrm{AM}}
=
\{
\theta_c
\}_{c\in \actioncases}.
$
The lifted \pam{} distributions $p_{\actionparams}(R_p \mid a)$ are applied on specific planning instances $I$ by assuming the roles of ground predicates are conditionally independent given the action $a$, i.e. the random vector \rolevar of roles for all ground predicates in $\predicates_I$ is governed by the distribution
\begin{align}
    \label{eq:role-distribution}
    p_{\actionparams}(\rolevar \mid a)
    =
    \textstyle\prod_{p \in \predicates_I}
    p_{\actionparams}(R_p \mid a),
\end{align}
where we slightly abuse notation by identifying the ground predicate $p \in \predicates_I$ with its corresponding predicate template.

In the \act{pick-up($x$)} example, $\mathit{holding}(x)$ has the \role{add} role, while $\mathit{onTable}(x)$ has the \role{del} role. Evidence from different ground instances \act{pick-up(A)}, \act{pick-up(B)}
therefore update the same lifted action-model parameters.
The probabilistic transition that maps role probabilities and a proposition values to a successor value is introduced with our model (Section~\ref{sec:nesyam}).


\paragraph{Variational inference.}
An alternative to maximising the log-likelihood $\log p_{\params}(\vec{O})$ of a series of observations \vec{O} under a parametrised distribution $p_{\params}(\vec{O}, \vec{X}) = p_{\params}(\vec{O} \mid \vec{X})\cdot p_{\params}(\vec{X})$ is to aim for the posterior distribution $p_{\params}(\vec{X} \mid \vec{O})$.
Indeed, the posterior is interpreted as the distribution over \vec{X} that is consistent with the given observations \vec{O}, instead of the distribution $p_{\params}(\vec{X})$ optimised to fit \vec{O}.
It is however generally intractable to compute $p_{\params}(\vec{X} \mid \vec{O})$.
Hence, variational inference (VI)~\cite{jordan1999introduction} approximates $p_{\params}(\vec{X} \mid \vec{O})$ by minimising a statistical divergence between $p_{\params}(\vec{X} \mid \vec{O})$ and a parametrised family of \emph{variational distributions} $q_{\viparams}(\vec{X})$.
Minimising the KL divergence is a popular choice in VI and is equivalent to maximising the \emph{evidence lower bound} (ELBO)~\cite{blei2017variational}
\begin{align}
    \label{eq:elbo}
    \expectation{
        q_{\viparams}(\vec{X})
    }{
        \log p_{\params}(\vec{O} \mid \vec{X})
    }
    -
    \kl{
        q_{\viparams}(\vec{X})
    }{
        p_{\params}(\vec{X})
    }
    .
\end{align}
The ELBO is popular, because choosing a fitting family of variational distributions, e.g. the mean-field family
\begin{align}
    q_{\viparams}(\vec{X})
    =
    \textstyle\prod_{i=1}^N
    q_{\viparams}(X_i),
\end{align}
can make it (partially) analytical.
Another reason is the easy fitting of any variational distribution using black-box variational inference~\cite{ranganath2014black}.


\section{Action-Model Learning under Partial Observability}
\label{sec:po-action-model-learning}
We now formalise the problem of learning action models from partially observable visual traces.
Let $I$ be the concrete planning instance of a lifted planning domain
$
\langle \types, \predicates, \actions,\mathrm{AM}\rangle
$.
Given visual traces of images $\emissionvec = (z_1, \dots, z_T)$,
grounded actions $\actionvec = (a_1, \dots, a_T) \in \actions_I^T$ and a
final states $f \in \predicates_I$,
the learning target is the action model $\actionmodel(a(\vec{x}))$ for every action schema $a(\vec{x})\in \actions$
while accounting for uncertainty about the hidden symbolic variables \statevec.
That is, a partially observable action model learning problem defines a parametrised probabilistic model
$
p_{\params}(\statevar, \emissionvec_{2:T}, f \mid z_1, \actionvec)
$
over symbolic states and images that takes the shape of the hidden Markov model (Figure~\ref{fig:hmm-model})
\begin{align*}
    p_{\params}(S_1 \mid z_1)
    \prod_{t=1}^{T}
    p_{\params_{\actionmodel}}\left(
        S_{t+1}\mid S_t,a_t
    \right)
    p_{\params}\left(
        z_t \mid S_t
    \right)
    \indicator{f = S_{T}},
\end{align*}
where we explicitly indicate the dependency of the transitions
$
p_{\params_{\mathrm{AM}}}\left(
        S_{t+1}\mid S_t,a_t
\right)
$
on the action model parameters $\params_{\actionmodel} \subseteq \params$.
We consider a joint probability distribution conditional on the first image $z_1$ to allow the distribution of the initial state $S_1$ to be informed by the first image $z_1$ through the perception model $p(S_1 \mid z_1)$.
One can remove the perception model and consider only emission models from states $S_t$ to $z_t$ via $p(z_t \mid S_t)$ in Figure~\ref{fig:hmm-model}.
The subsequent factorisation will instead start from an unconditional prior belief $p(S_1)$ that has to be obtained by other means.

\input{AAAI27/TIKZ/hmm-model}



The difficulty of action model learning under partial observability is compounded by the need to infer the symbolic states from images while the action model is itself unknown.
That is, there are two sources of uncertainty corresponding to different factors of $p(\statevar, \emissionvec_{2:T}, f \mid z_1, \actionvec)$;
the uncertainty on the action model is captured by the transitions
\textcolor{celadon_blue}{$p_{\params_{\actionmodel}}(S_{t + 1} \mid S_t, a_t)$},
while the uncertainty due to partial observability is captured by the perception model
\textcolor{red_salsa}{$p_{\params}(S_1 \mid z_1)$}
and emission model
\textcolor{jungle_green}{$p_{\params}(z_t \mid S_t)$}.
For example, if the gripper is hidden after \act{pick-up(A)}, the successor image may not reveal whether $\mathit{holding}(A)$ is true.
Nevertheless, the proposition $\mathit{holding}(A)$ may be constrained by observations, executed actions, or the final-state description.
The relevant question is therefore not only whether a proposition is visible in an individual image, but whether its value is identifiable from the complete trace.

Now let
$
\data
=
\left\{
\left(
\emissionvec^{(i)},
\actionvec^{(i)},
f^{(i)}
\right)
\right\}_{i=1}^{N}
$
be a dataset of visual traces of a planning instance $I$.
The action model \actionmodel as well as the perception model
$
p_{\params}(S_1 \mid z_1)
$
and emission model
$
p_{\params}(z_i \mid S_i)
$
can be jointly learned from \data by maximising the conditional log-likelihood of the dataset \data, i.e.
{\fontsize{9}{10.8}\selectfont
\begin{align}
    \label{eq:data-likelihood}
    \log p_{\params}(\data \mid Z_1, \actionvar)
    =
    \textstyle\sum_{i = 1}^N
    \log p_{\params}(\emissionvec_{2:T}^{(i)}, f^{(i)} \mid z_1^{(i)}, \actionvec^{(i)}).
\end{align}}%
Because the symbolic states are hidden, the likelihood
$
p_{\params}(
    \emissionvec_{2:T}^{(i)}, f^{(i)}
    \mid
    z_1^{(i)}, \actionvec^{(i)}
)
$
of a single data point requires a marginalisation over ground symbolic states
\begin{align}
    \label{eq:trace-marginal}
    &
    \textstyle\sum_{\statevec \in 2^{\predicates_I^T}}
    p_{\params}(
        \statevec, \emissionvec_{2:T}^{(i)}, f^{(i)}
        \mid
        z_1^{(i)}, \actionvec^{(i)}
    )
    \nonumber
    \\
    =\
    &\expectation{
        p_{\params}(
            \statevar
            \mid
            z_1^{(i)}, \actionvec^{(i)}
        )
    }{
        p_{\params}(
            \emissionvec_{2:T}^{(i)}, f^{(i)}
            \mid
            \statevar,
            z_1^{(i)}, \actionvec^{(i)}
        )
    }
\end{align}
This marginalisation is generally intractable as the state space $2^{\predicates_I}$ of even a single symbolic state $\statevec_i$ is combinatorially large and the data likelihood 
$
p_{\params}(
    \emissionvec_{2:T}^{(i)}, f^{(i)}
    \mid
    \statevar,
    z_1^{(i)}, \actionvec^{(i)}
)
$
depends on the full symbolic state $\statevar_t$.
The intractability is the reason for introducing approximations.






\section{Action-Model Learning with Relational Neurosymbolic Markov Models}
\label{sec:nesyam}
At a high-level, the probabilistic graphical model of action model learning (Figure~\ref{fig:hmm-model}) is directly captured by \nesymm{}s as it is a Markov model over symbolic states with subsymbolic observations.
The crucial point lies in how to model the transitions
$
p_{\params_{\actionmodel}}\left(
    S_{t+1}\mid S_t,a_t
\right)
$.
Here, we incorporate the probabilistic action model (\pam{}) of \citet{xi2024neuro} into a \nesymm{} according to STRIPS.



\paragraph{Action-Conditioned probabilistic transitions}
Given a symbolic state $s_t \in \predicates_I$ and an input ground action $a_t$, the \pam{} module outputs the joint distribution $p_{\actionparams}(\rolevar_t \mid a_t)$ over roles $\rolevar_t$ of every ground predicate $p \in \predicates_I$ (Equation~\ref{eq:role-distribution}) at time $t$. 
It specifies for every predicate what the probability is of that predicate taking a certain role with respect to the action $a_t$. 
For instance, the predicate $holding(A)$ should have a high probability for the role $\role{add}$ given the action $\act{pick-up(A)}$.

The goal of this section is to describe how to exactly compute the action-conditioned probabilistic transitions
$
p_{\actionparams}(S_{t + 1} \mid s_t, a_t)
$
that follow from combining the \pam{} predictions $p_{\actionparams}(\rolevar \mid a_ts)$ with the semantics of STRIPS.
The main difficulty lies in enforcing that the semantics of STRIPS are not violated.
For instance, not every role $r_{t, p}$ can be applied to every current state $s_{t, p}$.
The combination $r_{t, p} = \role{del}$ and $s_{t, p} = 0$ (false) is not allowed since the role \role{del} means that the predicate is both a precondition and it is deleted;
if $s_{t, p}$ is false, it can not be a precondition.
To make this modeling detail explicit, let $\varphi$ be the constraint
\begin{align}
    &\neg (s_{t, p} = 0 \land r_{t, p} = \role{pre})
    \\
    \land
    &\neg (s_{t, p} = 0 \land r_{t, p} = \role{del})
    \\
    \land
    &\neg (s_{t, p} = 1 \land r_{t, p} = \role{add})
\end{align}
imposed by STRIPS on every tuple $(r_{t, p}, s_{t, p})$.
The first line says that predicate $p$ can not be a prevail condition \role{pre} if it is currently false.
The second expresses that $p$ has to be true to be a deleted precondition \role{del}.
Both follow from the STRIPS semantics and the definitions of \role{pre} and \role{del}.
The third adds the non-standard constraint that true predicates can not be added by \role{add}.
The latter is introduced on top of the usual STRIPS constraints in order to avoid reasoning shortcuts~\cite{marconato2023not}, which was tackled by~\citet{xi2024neuro} by adding an additional losses.
Here, we build the constraint into the transitions.
$
p(S_{t + 1} \mid s_t, a_t, \varphi).
$
To do so, we start by marginalising out the roles $\rolevec_t$
\begin{align}
    p(S_{t + 1} \mid s_t, a_t, \varphi)
    =
    \textstyle\sum_{\rolevec_t}
    p(S_{t + 1}, \rolevec_t \mid s_t, a_t, \varphi),
\end{align}
and apply Bayes' rule to rewrite each summand as
\begin{align}
    &p(S_{t + 1}, \rolevec_t \mid s_t, a_t, \varphi)
    \nonumber
    \\
    =
    &
    \frac{
        p(S_{t + 1}, \rolevec_t \mid s_t, a_t)
        p(\varphi \mid s_t, a_t, \rolevec_t, S_{t + 1})
    }{
        p(\varphi \mid s_t, a_t)
    }
    \\
    =
    &
    \frac{
        p(S_{t + 1}, \rolevec_t \mid s_t, a_t)
        p(\varphi \mid s_t, \rolevec_t)
    }{
        \sum_{s_{t + 1}}
        \sum_{\rolevec_t'}
        p(S_{t + 1}, \rolevec_t' \mid s_t, a_t)
        p(\varphi \mid s_t, \rolevec_t')
    }
    .
    \label{eq:simplified-bayes-next-role-joint}
\end{align}
The joint
$
p(S_{t + 1}, \rolevec_t \mid s_t, a_t)
$
factorises as
\begin{align}
    \label{eq:state-role}
    p(S_{t + 1} \mid \rolevec_t, s_t) 
    p(\rolevec_t \mid a_t)
\end{align}
since
$
(S_{t + 1} \independent a_t \mid \rolevec_t)
$
and
$
(\rolevec_t \independent s_t \mid a_t)
$.
Once a role $\rolevec_t$ is assigned to each predicate in the symbolic state $s_t$ for the action $a_t$, the next state transition $p(S_{t + 1} \mid \rolevec_t, s_t)$ is determined deterministically according to the STRIPS semantics.
To be precise, the role $r_{t, p}$ of each predicate $p \in \predicates_I$ only depends on and affects the predicate $p$, i.e. $\forall p \neq q \in \predicates_I: (s_{t + 1, p} \independent s_{t + 1, q} \mid r_{t, p}, s_{t, p})$,
such that
\begin{align}
    \label{eq:next-state-distribution}
    p(S_{t + 1} \mid r_{t, p}, s_t)
    =
    \textstyle\prod_{p \in \predicates_I}
    p(S_{t + 1, p} \mid r_{t, p}, s_{t, p}),
\end{align}
with $s_{t, p} \in \predicates_I$ being the component of the state $s_t$ corresponding to the predicate $p$ at time $t$.
If $r_{t, p}(s_{t, p}) \in \predicates_I$ denotes the successor state obtained by applying role $r_{t, p}$ to $s_{t, p}$,
then the unconditional transitions are deterministic functions, i.e.
$
p(S_{t + 1, p} \mid r_{t, p}, s_{t, p})
=
\indicator{S_{t + 1, p} = r_{t, p}(s_{t, p})}
$.
Together with the factorisation of $p(\rolevec_t \mid a_t)$ (Equation~\ref{eq:role-distribution}), Equation~\ref{eq:state-role} reduces to
\begin{align}
    &\textstyle\prod_{p \in \predicates_I}
    \indicator{S_{t + 1, p} = r_{t, p}(s_{t, p})}
    p_{\actionparams}(r_{t, p} \mid a_t).
\end{align}
The constraint factor $p(\varphi \mid s_t,\rolevec_t)$ in Equation~\ref{eq:simplified-bayes-next-role-joint} similarly factorises over predicates
\begin{align}
    p(\varphi \mid s_t, \rolevec_t)
    =
    \textstyle\prod_{p \in \predicates_I}
    \indicator{\varphi(s_{t, p}, r_{t, p})},
\end{align}
so that Equation~\ref{eq:simplified-bayes-next-role-joint} also transitions predicate-wise
\begin{align}
    &p(S_{t + 1} \mid s_t, a_t, \varphi)
    =
    \\
    &
    \prod_{\predicates_I}
    \frac{
        \sum_{r_{t, p}}
        \indicator{S_{t + 1, p} = r_{t, p}(s_{t, p})}
        \indicator{\varphi(s_{t, p}, r_{t, p})}
        p_{\actionparams}(r_{t, p} \mid a_t)
    }{  
        \sum_{s_{t + 1, p}, r_{t, p}'}
        \indicator{s_{t + 1, p} = r_{t, p}'(s_{t, p})}
        \indicator{\varphi(s_{t, p}, r_{t, p}')}
        p_{\actionparams}(r_{t, p}' \mid a_t)
    }
    \nonumber
\end{align}
In summary, the constrained transition function
$
p_{\actionparams}(S_{t + 1} \mid s_t, a_t, \varphi)
$
factorises over predicates
$
\prod_{p \in \predicates_I}
p_{\actionparams}(S_{t + 1, p} \mid s_{t, p}, a_{t, p}, \varphi)
$
where every factor can be explicitly written in terms of the \pam{} probabilities as
\begin{align}
    \label{eq:next-state-pam}
    &
    p_{\actionparams}^{S_{t + 1, p}}(1 \mid s_{t, p}, a_t, \varphi)
    \\
    &
    {=} 
    \begin{cases}
        p_{\role{a}} / (p_{\role{a}} + p_{\role{n}})
        \quad
        &\text{if }
        s_t = 0, \\
        (p_{\role{n}} + p_{\role{p}}) / (p_{\role{n}} + p_{\role{p}} + p_{\role{d}})
        \quad
        &\text{if }
        s_t = 1,
    \end{cases}
    \nonumber
\end{align}
for role probabilities $p_{\role{n}},  p_{\role{p}},  p_{\role{a}}$ and $p_{\role{d}}$ corresponding to $r_{t, p}$.

This discussion fully specifies how \nesyam{} models and computes the probabilistic transitions that arise when using a \pam{} together with the formal Boolean semantics of STRIPS.
It is one of the key factors that differentiate \nesyam{} from the state of the art that relies on a fuzzy approximation of STRIPS.
Another key differentiating factor of \nesyam{} is its sampling-based approximation outlined in the next section.

\paragraph{Approximate action model learning with sampling}
Although the action model transition $p_{\actionparams}(S_{t + 1} \mid s_t, a_t, \varphi)$ factorises over predicates, exact maximum likelihood maximisation (Equation~\ref{eq:data-likelihood}) remains intractable due to the emission model $p_{\params}(z_t \mid S_t)$ generally depending on the full state $S_t$.
Hence, we propose two ways to optimise \nesyam{}:
\begin{enumerate*}[label=\textbf{(\arabic*)}]
    \item approximate maximum likelihood and
    \item a variational approximation of the posterior $p_{\params}(\statevar \mid \emissionvec, f, \actionvec, \varphi)$.
\end{enumerate*}
Both approaches rely on \nesyam{}'s ability to sample differentiable symbolic trajectories from $p_{\params}(\statevar \mid z_1, \actionvec)$ inherited from the underlying \nesymm{} model~\cite{de2025relational}.
Intuitively, \nesyam{} samples trajectories from the full joint distribution 
$
p_{\params}(
    \statevar
    \mid
    z_1, \actionvec, \varphi
)
$
to infer and optimise Equation~\ref{eq:data-likelihood}.
It first draws $N$ initial state samples $s_1^{(n)}$ from the predicted belief $p_{\params}(S_1 \mid z_1)$
Then, it repeatedly computes $p_{\params}(S_{t + 1} \mid s_t^{(n)}, a_t, \varphi)$ (Equation~\ref{eq:next-state-pam}) and $s_{t + 1}^{(n)}$ until a full trajectory is sampled.
The resulting trajectories $\left\{\statevec^{(n)}\right\}_{n = 1}^N$ are used in a Monte Carlo approximation of the maximum likelihood (Equation~\ref{eq:data-likelihood}) with approximated gradients~\cite{kool2019buy} (supplementary material)

\paragraph{Variational action model learning.} Alternatively, we can substitute a variational family of approximations
$
q_{\viparams}(\statevar \mid z_1, \actionvec)
$
into the ELBO (Equation~\ref{eq:elbo}) to target the posterior $p_{\params}(\statevar \mid \emissionvec, f, \actionvec, \varphi)$.
The result is an ELBO for variational action model learning
\begin{align}
    \label{eq:am-elbo}
    \sum_{i = 1}^N\
    &\expectation{
        q_{\viparams}(\statevar \mid z_1^{(i)}, \actionvec^{(i)})
    }{
        \log p_{\params}(\emissionvec^{(i)}_{2:T}, f^{(i)}, \varphi \mid \statevar, z_1^{(i)})
    } \\
    -\
    &\kl{q_{\viparams}(\statevar \mid z_1^{(i)}, \actionvec^{(i)})}{p_{\params}(\statevar \mid z_1^{(i)}, \actionvec^{(i)}, \varphi)}
    \nonumber,
\end{align}
over a dataset of visual traces 
$
\data
=
\left\{
\left(
\emissionvec^{(i)},
\actionvec^{(i)},
f^{(i)}
\right)
\right\}_{i=1}^{N}
$.
The default choice of $q_{\viparams}(\statevar \mid z_1, \actionvec)$ made by \nesyam{} is to use the parametrised prior, i.e.
\begin{align}
    \label{eq:nesyam-vi-approximation}
    q_{\viparams}(\statevar \mid z_1, \actionvec)
    =
    p_{\params}(\statevar \mid z_1, \actionvec, \varphi),
\end{align}
which simplifies the ELBO to the expected log-likelihood
\begin{align}
    \label{eq:nesyam-elbo}
    \textstyle\sum_{i = 1}^N
    \expectation{
        p_{\params}(\statevar \mid z_1^{(i)}, \actionvec^{(i)}, \varphi)
    }{
        \log p_{\params}(\emissionvec^{(i)}_{2:T}, f^{(i)} \mid \statevar, z_1^{(i)} 
    }.
\end{align}
The variational perspective on action model learning (Equation~\ref{eq:am-elbo}) will be used to theoretically compare \nesyam{} to \rosamei{}, the state-of-the-art action model learner of~\citet{xi2024neuro}.

\section{Related work with a variational flavour}
\label{sec:vi-perspective}

Action-model learning has been studied in both stochastic and deterministic planning settings \cite{pasula2007learning,aineto2019learning}.
Early and classical approaches learn from plan examples, noisy traces, or action traces using weighted MAX-SAT, logical inference, or relational model acquisition \cite{yang2007learning,zhuo2010learning,zhuo2013action,cresswell2009acquisition,cresswell2013acquiring}.
More recent work studies stronger formal guarantees, partial observability, noisy observations, missing action parameters, and lifted partial traces \cite{le2024learning,lamanna2024action,aineto2024action,balyo2024planning,lamanna2025lifted}.
These methods operate primarily over symbolic traces, while our setting requires visual traces.

Planning from images and representation learning for planning have explored how latent or symbolic representations can support planning \cite{asai2022classical,konidaris2018skills}.
\citet{xi2026learning} address the problem of removing action supervision, with fully observed images, initial and final states, which is orthogonal to the partial state observability studied here.
Closer to our observability setting, \citet{jin2022learning} learn latent transition models from partially observed images without a declarative action model, 
and \citet{zhu2025psalm} induce action semantics with large language models in interactive, partially observed visual environments
.
None learn lifted probabilistic action models from offline visual traces with images that underdetermine the symbolic state.

Our probabilistic framing builds on hidden Markov models and probabilistic graphical models \cite{koller2009probabilistic}.
Neurosymbolic AI combines neural perception with symbolic structure \cite{de2016statistical,marra2024statistical}, which \nesymm{}s extend to sequential probabilistic models \cite{de2025relational}.
\nesyam{} adapts \nesymm{}s to the problem of action-model learning from visual traces.

The current state of the art in neurosymbolic learning from visual traces, \rosamei{}, approximates a mean-field variational approach to action model learning.
\rosamei{} relaxes the underlying model (Figure~\ref{fig:hmm-model}) in two immediate ways.
First, it uses a fuzzy relaxation of the unconstrained transition function $p_{\actionparams}(S_{t + 1} \mid S_t, a_t)$, while the true model, and our approximation, computes the exact constrained transition function $p_{\actionparams}(S_{t + 1} \mid S_t, a_t, \varphi)$.
Second, it removes the emission model $p_{\params}(z_t \mid S_t)$ and, consequently, the maximisation of the likelihood of the images \emissionvec.
Instead, all images are used as inputs to the perception model $p_{\params}(S_t \mid z_t)$ to form a ``transition'' loss function that compares the state transitions according to the relaxed $p_{\actionparams}(S_{t + 1} \mid S_t, a_t)$ with the perception's predictions.
A separate fuzzy ``applicability'' loss is then added to promote adherence to the STRIPS constraints $\varphi$, as well as a supervision loss for the final state $s_T$.
While \rosamei{} does not frame itself as a variational approach, we can recover the foundational quantities underlying the losses by substituting the mean-field approximation
$
q_{\viparams}(\statevar)
=
\prod_{t = 1}^{T}
p_{\params}(S_t \mid z_t)
$
into the general action model learning ELBO (Equation~\ref{eq:am-elbo}) without emission likelihoods:
{\fontsize{6.5}{7.8}\selectfont
\begin{align}
    \label{eq:rosame-elbo}
    \sum_{i = 1}^N\
    &\expectation{
        p_{\params}(S_{T} \mid z_{T}^{(i)})
    }{
        \log p_{\params}(f^{(i)} \mid S_T, z_1^{(i)} 
    } 
    +
    \expectation{
        q_{\viparams}(\statevar)
    }{
        \log p(\varphi \mid \statevar)
    }\
    - \\
    &
    \sum_{t = 1}^{T - 1}
    \expectation{
        p_{\params}(S_{t} \mid z_{t}^{(i)})
    }{
        \kl{p_{\params}(S_{t + 1} \mid z_{t + 1}^{(i)})}{p_{\actionparams}(S_{t + 1} \mid s_{t})}
    },
    \nonumber
\end{align}}%
where we have left out the dependency of $p_{\params}(S_{t + 1} \mid S_{t})$ on $z_1^{(i)}$ and $\actionvec^{(i)}$ on the final line for brevity.
The three terms provide theoretical justification for the supervision loss, applicability loss and transition loss, respectively.
It also shows why \rosamei{} is not expected to work well in partial observable environments;
if the images $z_i$ are incomplete or noisy,
then each pairwise term in the transition loss only enforces local consistency on the noisy predictions from $p_{\params}(S_t \mid z_t)$.
In contrast, \nesyam{} approximations explicitly model the sequential dependencies between the states $S_t$ and aim to incorporate the observed images \emissionvec and final state $f$ (Equation~\ref{eq:nesyam-elbo}).
Hence, it can remain coherent, albeit more uncertain, if some images $z_t$ are not observed or only contains partial information about the state $S_t$.
This perspective gives a theoretical reason why \nesyam{} improves upon \rosamei{} under partial observability that we empirically confirm next.



\section{Experiments}

\begin{figure}[t]
  \centering
  \includegraphics[width=0.45\textwidth]{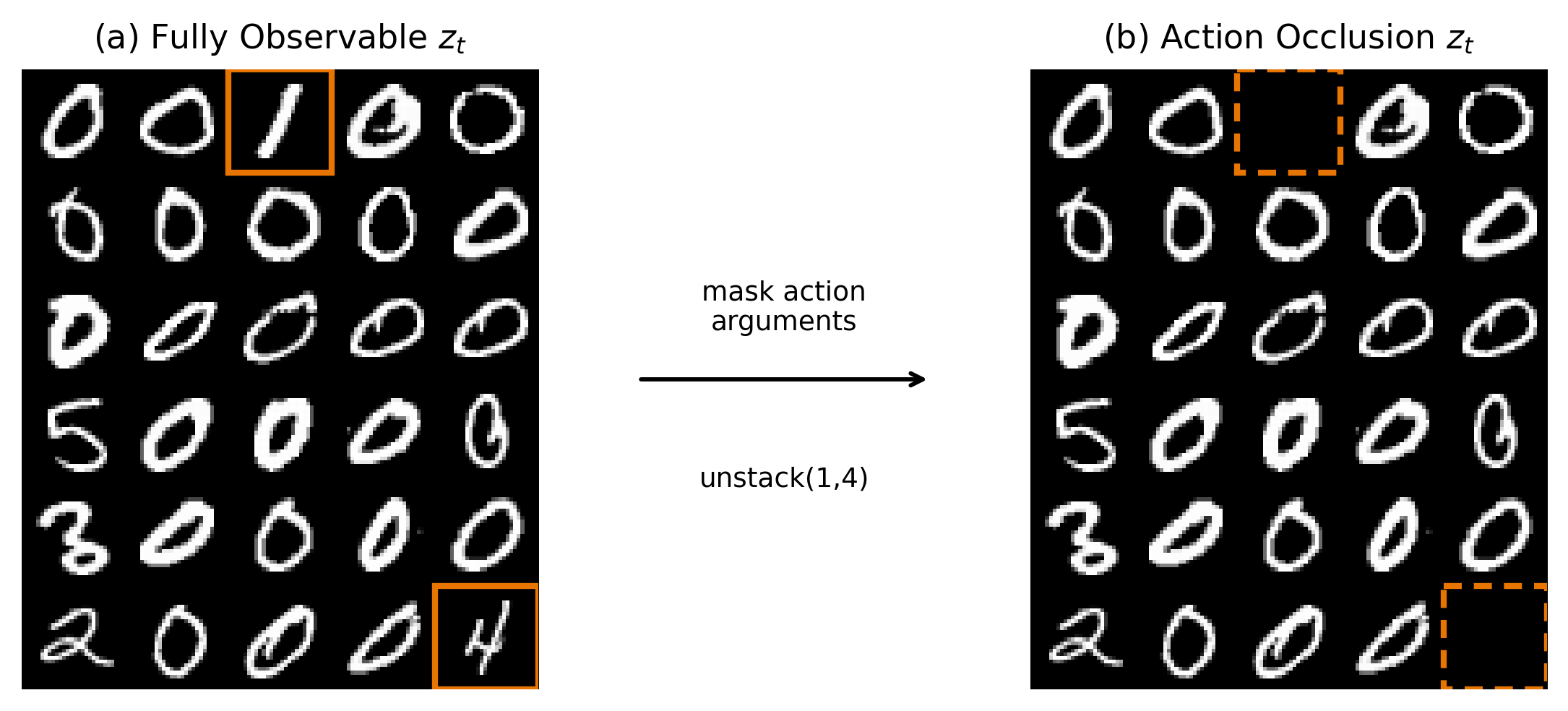}
    \caption{Action Occlusion in grid-based Blocks World. For \act{unstack(1,4)}, solid boxes mark the action arguments in the Fully observable case~\ref{o1} (left), while dashed boxes mark their masked locations under Action occlusion~\ref{o3} (right).}
  \label{fig:action-occlusion}
\end{figure}

\input{AAAI27/Results/A1}

\label{sec:experiments}
Our experiments aim to answer the following questions.
\begin{enumerate}[
    label=\textbf{(Q\arabic*)},
    ref=Q\arabic*,
    leftmargin=0pt,
    labelsep=0.5em,
    labelwidth=0pt,
    itemindent=*,
    align=left,
    itemsep=0.25em
]
    \item How do \nesyam{} and \rosamei{} compare in lifted action-model recovery under partial observability?
    \item How does learning \nesyam{} and \rosamei{} scale as the grounded planning problem grows?
    \item How does the number of sampled trajectories affect \nesyam{}'s action-model recovery and training time?
\end{enumerate}

\paragraph{Domains, data and modeling}
Following the setup of~\citet{xi2024neuro}, we evaluate both methods on six visual planning domains: grid-based Blocks World (720/80 training/test traces), Gripper (900/100), Logistics (2250/250), synthesized-image Blocks World (100/100), Towers of Hanoi (70/100), and 8-Puzzle (300/100).
Traces contain ten actions in both Blocks World domains and Logistics, and five actions otherwise.
Both \nesyam{} and \rosamei{} use identical \pam{} modules and perception models, while \nesyam{} uses a neural emission model (supplementary material).
\nesyam{} maximises the approximate maximum likelihood objective and \rosamei{} utilises its originally proposed losses (hyperparameters in supplementary material).

\paragraph{Observation regimes}
We evaluate each domain under three observation regimes.
\begin{enumerate*}[
    label=\textbf{(O\arabic*)},
]
    \item \emph{Fully observable} leaves all observations unchanged,\label{o1}
    \item \emph{Random occlusion} randomly masks selected regions of each non-initial observation and\label{o2}
    \item \emph{Action-occlusion} masks action-relevant regions based on the executed action or its visual effect\label{o3} (Figure~\ref{fig:action-occlusion}).
\end{enumerate*}
Both occlusion regimes keep maintain fully observed initial observations and final-state descriptions.
\paragraph{Action-model evaluation and evaluation scope}
For each action schema, we test whether a type-compatible predicate template is assigned its correct role
by comparing the role with highest probability to the ground truth role.
Two complementary scores are reported corresponding to different subsets of action-predicate pairs of interest.
On the one hand, the full-candidate subset $\mathcal{C}$ includes every type-compatible action-predicate pair and therefore penalizes false positives on predicates not involved in an action.
On the other hand, the relevant-role subset $\mathcal{C}_{\mathrm{rel}}$ includes only pairs whose reference role is \role{pre}, \role{add}, or \role{del}.
It measures recovery of the predicates defining an action but ignores false positives among action-predicate pairs with \role{not-involved} as reference role.


\paragraph{A1: \nesyam{} is robust to partial observations}
Table~\ref{tab:cross-domain-evaluation} shows action model recovery across all domains and observations regimes.
Each entry reports full-candidate followed by relevant-role recovery, with cross-domain totals in the final rows. 
\nesyam{} recovers $90.00$ of $90$ relevant roles in the fully observable case~\ref{o1}, $88.33$ under Random occlusion~\ref{o2}, and $89.33$ under Action occlusion~\ref{o3}.
\rosamei{} instead declines from $87.00$ to $61.33$ and $68.00$, respectively.
Hence, \nesyam{} outperforms \rosamei{} in every domain under both occlusion regimes in terms of relevant-role predicates. 
Both methods recover slightly more relevant roles under Action occlusion~\ref{o3} than Random occlusion~\ref{o2}, likely because the action-occluded masks reveal the action arguments and provide indirect supervision.
Conversely, full-candidate recovery yields a mixed comparison.
\nesyam{} delivers relatively consistent scores across observation regimes and performs better under Random occlusion~\ref{o2}, but \rosamei{} outperforms \nesyam{} under full observability~\ref{o1} and Action occlusion~\ref{o3}.
In conclusion, \nesyam{} performs significantly better on relevant role recovery under partial observability, but can be less accurate on \role{not-involved} roles.

\paragraph{A2: \nesyam{} maintains action-model recovery as grounding grows}
We evaluate \nesyam{} and \rosamei{} on Blocks World with an increasing number of blocks from five to eight under full observability~\ref{o1} (Table~\ref{tab:scalability}).
As the number of blocks increases, the number of ground atoms, e.g. $\mathrm{clear}(1)$ and $\mathrm{on}(1,2)$, grows from $36$ to $81$.
The number of ground actions grows from $50$ to $128$, while the action model remains unchanged as it is lifted.
While \rosamei{} performs better when training at the default number of five blocks on full candidate recovery, \nesyam{} provides stable full candidate recovery across all block numbers that surpasses \rosamei{}'s recovery from 7 blocks onwards.
In fact, \rosamei{} degrades significantly in both full candidate and relevant-role recovery as the number of blocks increases.
The decrease is mainly driven by an increasing number of mistakes on relevant-role recovery. 
In contrast, \nesyam{}'s maintains fully correct relevant-role predictions across all domain sizes, as well as maintain its performance on full candidate predictions.
\nesyam{} thus better preserves action-model recovery as grounding grows, but its training time increases more sharply from $1887.94$ to $3352.89$ seconds, compared to $1473.62$ to $1856.98$ seconds for \rosamei{}.

\input{AAAI27/Results/A2}


\paragraph{A3: Sampling more trajectories increases cost without improving recovery}
We vary the training sampling budget $N\in\{2,8,32,128,256\}$ for \nesyam{} on five-block Blocks World under Action Occlusion~\ref{o3}. 
Evaluation does use $256$ samples, keeping all other settings fixed. 
Table~\ref{tab:rollout-sensitivity} shows that \nesyam{} recovers all $18$ relevant roles for every $N$, while full-candidate recovery remains between $19.33$ and $20.00$.
True-role probability also shows no consistent improvement, ranging from $67.01\%$ to $69.85\%$. 
Interestingly, $N=2$ achieves the highest mean recovery and true-role probability with the lowest training time. 
Training time remains similar through $N=32$ but rises to $1963.48$ seconds at $N=128$ and $2951.03$ seconds at $N=256$. 
Thus, two samples are sufficient for learning in this setting and larger budgets increase cost without improving recovery.

\input{AAAI27/Results/A3}

\paragraph{Discussion}
The results show that \nesyam{} and \rosamei{} have different error profiles.
\nesyam{} recovers nearly all relevant relations and remains stable under both occlusion regimes, whereas \rosamei{} is more accurate on \role{not-involved} roles.
\nesyam{}'s advantage should therefore be understood as stronger preservation of action-defining predicates rather than uniformly better action-model recovery. 
\rosamei{}'s relatively strong performance under Action occlusion may partly arise because the action-occluded masks reveal action arguments and provide indirect supervision.

\section{Limitations}

\nesyam{}'s errors are mainly given by false positives among \role{not-involved} cases, which reflects an identifiability limitation of learning from positive traces.
Positive traces show which predicates hold when an action was successfully executed, but not which predicates were required. 
If a predicate is true before every execution and remains true afterward, both \role{pre} and \role{not-involved} explain the transition equally well.
Without additional evidence or prior preference, these roles cannot be distinguished and so form a so-called reasoning shortcut~\cite{marconato2023not}.
\rosamei{} resolves the identifiablity problem using a loss encoding an explicit bias for \role{pre} conditions, whereas \nesyam{} deliberately relies only on trace evidence.
False-positive preconditions can make the learned action model overly restrictive, while false-positive effects can produce incorrect successor states.
Future work could address this ambiguity through a minimality assumption, i.e. a prior preference, towards \role{not-involved} or through traces in which uninvolved predicates are sometimes false when the action is executed.
Such traces would allow the observations to distinguish the two roles.

\section{Conclusion}
This paper presents \nesyam{}, a probabilistic solution to the action model learning problem.
\nesyam{} integrates probabilistic predicate effects for any given action with the semantics of STRIPS into a relational neurosymbolic Markov model.
It uses end-to-end differentiable sampling for either maximum likelihood or variational inference approximations to learn action models.
The samples of \nesyam{} represent fully consistent trajectories over time that adhere to the rules of STRIPS.
This consistency over time is the main reason why \nesyam{} can deal with partial observability where the state of the art struggles, as shown theoretically by viewing action model learning through the lens of variational inference.
Empirically, across six domains and three observation regimes, \nesyam{} preserves relevant-role recovery more consistently than \rosamei{} under occlusion, although false positives on \role{not-involved} roles remain a limitation.

\bibliography{action_model_vi_2027}

\end{document}

%% file: macros.tex
\usepackage{amsthm}
\usepackage[inline]{enumitem}
\usepackage{booktabs}
\usepackage{tikz}
\usepackage{xspace}
\usepackage{circuitikz}
\usepackage{dsfont}
\usetikzlibrary{arrows.meta}
\usetikzlibrary{patterns.meta}
\usetikzlibrary{calc,shapes,backgrounds,fit,arrows,positioning,decorations.pathmorphing,patterns}

\usepackage{sourcecodepro}

\definecolor{red_salsa}{HTML}{F94144}
\definecolor{orange_red}{HTML}{F3722C}
\definecolor{yellow_orange}{HTML}{F8961E}
\definecolor{mango_tango}{HTML}{F9844A}
\definecolor{maize_crayola}{HTML}{F9C74F}
\definecolor{pistachio}{HTML}{90BE6D}
\definecolor{jungle_green}{HTML}{43AA8B}
\definecolor{steel_teal}{HTML}{4D908E}
\definecolor{queen_blue}{HTML}{577590}
\definecolor{celadon_blue}{HTML}{277DA1}
\definecolor{softer_black}{HTML}{2A2A2A}

\definecolor{poinant_purple}{HTML}{A87DC1}
\definecolor{purpose_purple}{HTML}{4D5AC0}
\definecolor{bliss_brown}{HTML}{C36D3A}
\definecolor{turd_turqoise}{HTML}{44BAC3}

\definecolor{soft_white}{HTML}{F1F1F1}
\definecolor{softer_white}{HTML}{E5E5E5}
\definecolor{grey}{HTML}{A5A5A5}
\definecolor{softer_grey}{HTML}{D1D1D1}
\definecolor{soft_grey}{HTML}{C7C7C7}
\definecolor{dark_grey}{HTML}{8E8E8E}
\definecolor{soft_black}{HTML}{101010}
\definecolor{softer_black}{HTML}{2A2A2A}

\tikzset{
  circnode/.style={
    fill=softer_white, align=center,
    circle, 
    minimum width=2.5em
  },
  sqnode/.style={
    fill=softer_white,
    rounded corners=0.2em,align=center,
    minimum width=2.5em, minimum height=2.5em
  },
  diamnode/.style={
    fill=softer_white,
    diamond, 
    shape aspect=1,
    rounded corners=0.2em,align=center,
    minimum width=3.2em, minimum height=3.2em
  },
  regpath/.style={
    -Triangle[round], draw=softer_black, ultra thick, rounded corners=0.5em
  },
  thinpath/.style={
    -Triangle[round], draw=softer_black, thick, rounded corners=0.5em
  },
  thinpathdouble/.style={
    Triangle[round]-Triangle[round], draw=softer_black, thick, rounded corners=0.5em
  },
  decision/.style={
    sqnode,
    fill=lavender_purple,
  },
  randvar/.style={
    circnode,
    fill=pale_blue,
  },
  obsvar/.style={
    circnode,
    fill=tea_green,
  },
  reward/.style={
    diamnode,
    fill=soft_yellow,
  },
  evidence/.style={
    circnode,
    fill=jungle_green!25
  }
}

\tikzset{
    figpgms/.style={
        inner sep=0mm,
        minimum width=10mm,
        minimum height=10mm
    }
}

\tikzdeclarepattern{
  name=custom_line,
  parameters={
      \pgfkeysvalueof{/pgf/pattern keys/size},
      \pgfkeysvalueof{/pgf/pattern keys/angle},
      \pgfkeysvalueof{/pgf/pattern keys/line width},
  },
  bounding box={
    (0,-0.5*\pgfkeysvalueof{/pgf/pattern keys/line width}) and
    (\pgfkeysvalueof{/pgf/pattern keys/size},
0.5*\pgfkeysvalueof{/pgf/pattern keys/line width})},
  tile size={(\pgfkeysvalueof{/pgf/pattern keys/size},
\pgfkeysvalueof{/pgf/pattern keys/size})},
  tile transformation={rotate=\pgfkeysvalueof{/pgf/pattern keys/angle}},
  defaults={
    size/.initial=5pt,
    angle/.initial=90,
    line width/.initial=7pt,
    xshift/.initial=1.1*\n cm,
    distance/.initial=3pt,
  },
  code={
      \draw [line width=\pgfkeysvalueof{/pgf/pattern keys/line width}]
        (0,0) -- (\pgfkeysvalueof{/pgf/pattern keys/size},0);
  },
}

\theoremstyle{definition}

\theoremstyle{definition}

\theoremstyle{definition}

\theoremstyle{definition}

\theoremstyle{definition}

\theoremstyle{definition}

\newcommand{\actionmodel}{\ensuremath{\mathrm{AM}}\xspace}

\newcommand{\types}{\ensuremath{\mathcal{T}}\xspace}
\newcommand{\predicates}{\ensuremath{\mathcal{P}}\xspace}
\newcommand{\actions}{\ensuremath{\mathcal{A}}\xspace}
\newcommand{\objects}{\ensuremath{\mathcal{O}}\xspace}
\newcommand{\data}{\ensuremath{\mathcal{D}}\xspace}
\newcommand{\actioncases}{\ensuremath{\mathcal{C}}\xspace}

\renewcommand{\vec}[1]{\ensuremath{\boldsymbol{#1}}\xspace}

\newcommand{\statevec}{\vec{s}}
\newcommand{\emissionvec}{\vec{z}}
\newcommand{\actionvec}{\vec{a}}
\newcommand{\rolevec}{\vec{r}}

\newcommand{\statevar}{\vec{S}}

\newcommand{\actionvar}{\vec{A}}
\newcommand{\rolevar}{\vec{R}}

\newcommand{\params}{\ensuremath{\boldsymbol{\theta}}\xspace}
\newcommand{\actionparams}{\ensuremath{\boldsymbol{\theta}_{\actionmodel}}\xspace}
\newcommand{\viparams}{\ensuremath{\boldsymbol{\lambda}}\xspace}

\newcommand{\indicator}[1]{\mathds{1}_{#1}}
\newcommand{\expectation}[2]{\ensuremath{\mathbb{E}_{#1}\left[#2\right]}}
\newcommand{\kl}[2]{\ensuremath{\textsc{KL}\left(#1 \mid\mid #2\right)}}

\newcommand{\independent}{\perp\!\!\!\perp}

%% file: AAAI27/TIKZ/hmm-model.tex
\begin{figure}[t]
  \centering
\def\hmmnodesize{7mm}
\def\hmmnodepadding{0.8pt}
\def\hmmxgap{1.35}
\def\hmmygap{1.10}

\begin{tikzpicture}[
  font=\scriptsize,
  compactnode/.style={
    minimum size=\hmmnodesize,
    inner sep=\hmmnodepadding
  },
  circnode/.append style={compactnode},
  evidence/.append style={compactnode}
]

  \node[circnode] (s1) at (0,0)                 {$S_1$};
  \node[circnode] (s2) at (\hmmxgap,0)          {$S_2$};
  \node[circnode] (s3) at ({2*\hmmxgap},0)      {$S_3$};
  \node[circnode] (s4) at ({3*\hmmxgap},0)      {$S_4$};
  \node[circnode] (s5) at ({4*\hmmxgap},0)      {$S_5$};

  \node[evidence] (a1) at (0,\hmmygap)                 {$a_1$};
  \node[evidence] (a2) at (\hmmxgap,\hmmygap)          {$a_2$};
  \node[evidence] (a3) at ({2*\hmmxgap},\hmmygap)      {$a_3$};
  \node[evidence] (a4) at ({3*\hmmxgap},\hmmygap)      {$a_4$};

  \node[evidence] (z1) at (0,{-\hmmygap})                 {$z_1$};
  \node[evidence] (z2) at (\hmmxgap,{-\hmmygap})          {$z_2$};
  \node[evidence] (z3) at ({2*\hmmxgap},{-\hmmygap})      {$z_3$};
  \node[evidence] (z4) at ({3*\hmmxgap},{-\hmmygap})      {$z_4$};
  \node[evidence] (f)  at ({4*\hmmxgap},{-\hmmygap})      {$f$};


    \draw[thinpath, celadon_blue]
      (s1) --
      node[below=1mm] {$\mathrm{AM}$}
      (s2);
    \draw[thinpath, celadon_blue] (s2) -- (s3);
    \draw[thinpath, celadon_blue] (s3) -- (s4);
    \draw[thinpath, celadon_blue] (s4) -- (s5);

    \draw[thinpath, celadon_blue] (a1) -- (s2);
    \draw[thinpath, celadon_blue] (a2) -- (s3);
    \draw[thinpath, celadon_blue] (a3) -- (s4);
    \draw[thinpath, celadon_blue] (a4) -- (s5);

    \draw[thinpath, red_salsa]
      (z1) --
      node[left] {}
      (s1);

    \draw[thinpath, jungle_green]
      (s2) --
      node[right] {}
      (z2);
    \draw[thinpath, jungle_green] (s3) -- (z3);
    \draw[thinpath, jungle_green] (s4) -- (z4);

    \draw[thinpath] (s5) -- (f);

  \end{tikzpicture}

  \caption{
    Action-conditioned hidden Markov model for a visual trace of length
    $T=5$ representing the overall model to learn.
    Grey nodes indicate hidden symbolic state variables and \textcolor{jungle_green}{green} nodes
    indicate input actions and images.
    The perception model \textcolor{red_salsa}{$p_{\params}(S_1 \mid z_1)$} predicts the belief of the initial state.
    The probabilistic action model $\actionmodel$ and the observed grounded actions \actionvec produce the
    successor states through \textcolor{celadon_blue}{$p_{\params_{\actionmodel}}(S_{t + 1} \mid S_t, a_t)$}.
    The emission model \textcolor{jungle_green}{$p_{\params}(z_t \mid S_t)$} measures the agreement between the states and their associated images, while the final-state supervision
    compares the sampled endpoint $s_5$ with the observed description
    $f$.
  }
  \label{fig:hmm-model}
\end{figure}

%% file: AAAI27/Results/A1.tex
\begin{table*}[t]
    \centering
    \setlength{\tabcolsep}{1mm}
    \small
    \begin{tabular}{@{}llcrrr@{}}
        \toprule
        Domain & Method
        & $|\mathcal{C}|/|\mathcal{C}_{\mathrm{rel}}|$
        & Fully observed (All/Rel.)
        & Random occlusion (All/Rel.)
        & Action occlusion (All/Rel.) \\
        \midrule

        Blocks World
        & \nesyam{} & $26/18$
        & $19.33 \pm 0.58 \,/\, \mathbf{18.00 \pm 0.00}$
        & $\mathbf{20.00 \pm 0.00} \,/\, \mathbf{18.00 \pm 0.00}$
        & $\mathbf{19.67 \pm 0.58} \,/\, \mathbf{18.00 \pm 0.00}$ \\
        (grid world)
        & \rosamei{} & $26/18$
        & $\mathbf{26.00 \pm 0.00} \,/\, \mathbf{18.00 \pm 0.00}$
        & $15.67 \pm 3.06 \,/\, 7.67 \pm 3.06$
        & $18.33 \pm 2.89 \,/\, 10.33 \pm 2.89$ \\
        \addlinespace

        Gripper
        & \nesyam{} & $10/10$
        & $\mathbf{10.00 \pm 0.00} \,/\, \mathbf{10.00 \pm 0.00}$
        & $\mathbf{10.00 \pm 0.00} \,/\, \mathbf{10.00 \pm 0.00}$
        & $\mathbf{10.00 \pm 0.00} \,/\, \mathbf{10.00 \pm 0.00}$ \\
        & \rosamei{} & $10/10$
        & $\mathbf{10.00 \pm 0.00} \,/\, \mathbf{10.00 \pm 0.00}$
        & $9.00 \pm 0.00 \,/\, 9.00 \pm 0.00$
        & $6.00 \pm 0.00 \,/\, 6.00 \pm 0.00$ \\
        \addlinespace

        Logistics
        & \nesyam{} & $18/18$
        & $\mathbf{18.00 \pm 0.00} \,/\, \mathbf{18.00 \pm 0.00}$
        & $\mathbf{17.00 \pm 1.73} \,/\, \mathbf{17.00 \pm 1.73}$
        & $\mathbf{17.33 \pm 0.58} \,/\, \mathbf{17.33 \pm 0.58}$ \\
        & \rosamei{} & $18/18$
        & $17.67 \pm 0.58 \,/\, 17.67 \pm 0.58$
        & $16.00 \pm 0.00 \,/\, 16.00 \pm 0.00$
        & $14.33 \pm 0.58 \,/\, 14.33 \pm 0.58$ \\
        \addlinespace

        Blocks World
        & \nesyam{} & $26/18$
        & $19.33 \pm 0.58 \,/\, \mathbf{18.00 \pm 0.00}$
        & $\mathbf{19.67 \pm 0.58} \,/\, \mathbf{18.00 \pm 0.00}$
        & $19.00 \pm 0.00 \,/\, \mathbf{18.00 \pm 0.00}$ \\
        (synthesized)
        & \rosamei{} & $26/18$
        & $\mathbf{26.00 \pm 0.00} \,/\, \mathbf{18.00 \pm 0.00}$
        & $17.00 \pm 4.00 \,/\, 9.00 \pm 4.00$
        & $\mathbf{22.67 \pm 5.77} \,/\, 14.67 \pm 5.77$ \\
        \addlinespace

        Hanoi
        & \nesyam{} & $15/6$
        & $8.33 \pm 0.58 \,/\, \mathbf{6.00 \pm 0.00}$
        & $8.00 \pm 0.00 \,/\, \mathbf{6.00 \pm 0.00}$
        & $7.67 \pm 0.58 \,/\, \mathbf{6.00 \pm 0.00}$ \\
        & \rosamei{} & $15/6$
        & $\mathbf{12.67 \pm 2.08} \,/\, 4.33 \pm 1.53$
        & $\mathbf{12.67 \pm 2.08} \,/\, 4.33 \pm 1.53$
        & $\mathbf{13.00 \pm 1.73} \,/\, 4.67 \pm 1.15$ \\
        \addlinespace

        8-Puzzle
        & \nesyam{} & $96/20$
        & $43.33 \pm 1.53 \,/\, \mathbf{20.00 \pm 0.00}$
        & $\mathbf{50.00 \pm 3.46} \,/\, \mathbf{19.33 \pm 1.15}$
        & $40.00 \pm 4.58 \,/\, \mathbf{20.00 \pm 0.00}$ \\
        & \rosamei{} & $96/20$
        & $\mathbf{91.00 \pm 1.73} \,/\, 19.00 \pm 1.73$
        & $37.33 \pm 9.24 \,/\, 15.33 \pm 1.15$
        & $\mathbf{52.67 \pm 4.04} \,/\, 18.00 \pm 3.46$ \\

        \midrule
        Total recovery
        & \nesyam{} & $191/90$
        & $118.33 \pm 1.53 \,/\, \mathbf{90.00 \pm 0.00}$
        & $\mathbf{124.67 \pm 2.52} \,/\, \mathbf{88.33 \pm 2.89}$
        & $113.67 \pm 3.21 \,/\, \mathbf{89.33 \pm 0.58}$ \\
        & \rosamei{} & $191/90$
        & $\mathbf{183.33 \pm 4.04} \,/\, 87.00 \pm 3.61$
        & $107.67 \pm 2.52 \,/\, 61.33 \pm 8.14$
        & $\mathbf{127.00 \pm 6.56} \,/\, 68.00 \pm 13.00$ \\
        \bottomrule
    \end{tabular}
    \caption{Combined lifted action-model evaluation across observation regimes. Each entry reports the mean number of correctly assigned roles $\pm$ standard deviation over three seeds. The candidate-count column reports $|\mathcal{C}|$ and $|\mathcal{C}_{\mathrm{rel}}|$.
    The highest mean is highlighted in bold for each evaluation scope with ties all highlighted.}
    \label{tab:cross-domain-evaluation}
\end{table*}

%% file: AAAI27/Results/A2.tex
\begin{table}[t]
    \centering
    \small
    \setlength{\tabcolsep}{1mm}

    \begin{tabular}{@{}clcc@{}}
        \toprule
        Size
        & Method
        & \shortstack{Fully observed (All/Rel.)}
        & Time (s) \\
        \midrule

        5
        & \nesyam{}
        & $19.33 \pm 0.58 \,/\, \mathbf{18.00 \pm 0.00}$
        & $1887.94$ \\
        & \rosamei{}
        & $\mathbf{26.00 \pm 0.00} \,/\, \mathbf{18.00 \pm 0.00}$
        & $\mathbf{1473.62}$ \\
        \addlinespace

        6
        & \nesyam{}
        & $\mathbf{19.33 \pm 0.58} \,/\, \mathbf{18.00 \pm 0.00}$
        & $2289.39$ \\
        & \rosamei{}
        & $19.00 \pm 0.00 \,/\, 11.00 \pm 0.00$
        &  $\mathbf{1499.36}$ \\
        \addlinespace

        7
        & \nesyam{}
        & $\mathbf{19.67 \pm 0.58} \,/\, \mathbf{18.00 \pm 0.00}$
        & $2817.42$ \\
        & \rosamei{}
        & $16.00 \pm 0.00 \,/\, 8.00 \pm 0.00$
        & $\mathbf{1711.72}$ \\
        \addlinespace

        8
        & \nesyam{}
        & $\mathbf{19.00 \pm 0.00} \,/\, \mathbf{18.00 \pm 0.00}$
        & $3352.89$\\
        & \rosamei{}
        & $15.67 \pm 0.58 \,/\, 7.67 \pm 0.58$ 
        & $\mathbf{1856.98}$ \\
        \bottomrule
    \end{tabular}
    \caption{Scalability in Blocks World under full observability.
    Entries report the mean number of correctly recovered roles $\pm$ standard deviation across three seeds
    Time is the mean training time in seconds across the same seeds.}
    \label{tab:scalability}
\end{table}

%% file: AAAI27/Results/A3.tex
\begin{table}[t]
    \centering
    \small
    \setlength{\tabcolsep}{1mm}

    \begin{tabular}{@{}rccc@{}}
        \toprule
        $N$
        & \shortstack{Recovery\\(All/Rel.)}
        & \shortstack{True-role\\probability (\%)}
        & Time (s) \\
        \midrule

        2
        & $\mathbf{20.00 \pm 0.00} \,/\, \mathbf{18.00 \pm 0.00}$
        & $\mathbf{69.85 \pm 0.75}$
        & $\mathbf{1180.18}$ \\

        8
        & $19.33 \pm 0.58 \,/\, 18.00 \pm 0.00$
        & $67.01 \pm 2.13$
        & $1212.52$ \\

        32
        & $19.67 \pm 0.58 \,/\, 18.00 \pm 0.00$
        & $68.38 \pm 2.30$
        & $1203.42$ \\

        128
        & $19.67 \pm 0.58 \,/\, 18.00 \pm 0.00$
        & $68.70 \pm 2.60$
        & $1963.48$ \\

        256
        & $19.33 \pm 0.58 \,/\, 18.00 \pm 0.00$
        & $67.70 \pm 2.29$
        & $2951.03$ \\
        \bottomrule
    \end{tabular}
    \caption{Effect of the number of sampled trajectories $K$ on \nesyam{} in five-block Blocks World under Action-Conditioned Occlusion. Recovery is the mean number of correctly recovered roles $\pm$ standard deviation across three seeds
    . 
    True-role probability is the mean probability assigned to the reference roles, and Time is the mean training time in seconds.
    }
    \label{tab:rollout-sensitivity}
\end{table}